\documentclass[peerreview]{IEEEphot}

\usepackage{amsmath}
\usepackage{amssymb}
\usepackage{color}
\usepackage{array}
\usepackage{booktabs}
\usepackage{multirow}
\usepackage[numbers, sort&compress]{natbib}
\usepackage[
 bookmarks=true,bookmarksnumbered=true,bookmarksopen=true,bookmarksopenlevel=1,
 breaklinks=false,pdfborder={0 0 0},pdfborderstyle={},backref=false,colorlinks=true]{hyperref}
\usepackage{t1enc}
\usepackage{subfig}

\begin{document}

\title{Light Source Point Cluster Selection Based Atmosphere Light Estimation}

\author{Wenbo~Zhang, Xiaorong~Hou}

\affil{School of Energy Science and Engineering, University of Electronic
Science and Technology of China, Chengdu 611731, China}

\maketitle
\markboth{IEEE Photonics Journal}{Volume Extreme Ultraviolet Holographic Imaging}

\begin{abstract}
Atmosphere light value is a highly critical parameter in defogging
algorithms that are based on an atmosphere scattering model. Any error
in atmosphere light value will produce a direct impact on the accuracy
of scattering computation and thus bring chromatic distortion to restored
images. To address this problem, this paper propose a method that
relies on clustering statistics to estimate atmosphere light value.
It starts by selecting in the original image some potential atmosphere
light source points, which are grouped into point clusters by means
of clustering technique. From these clusters, a number of clusters
containing candidate atmosphere light source points are selected,
the points are then analyzed statistically, and the cluster containing
the most candidate points is used for estimating atmosphere light
value. The mean brightness vector of the candidate atmosphere light
points in the chosen point cluster is taken as the estimate of atmosphere
light value, while their geometric center in the image is accepted
as the location of atmosphere light. Experimental results suggest
that this statistics clustering method produces more accurate atmosphere
brightness vectors and light source locations. This accuracy translates
to, from a subjective perspective, more natural defogging effect on
the one hand and to the improvement in various objective image quality
indicators on the other hand.
\end{abstract}


\section{Introduction}

Image defogging has become one hot spot in the domain of outdoor surveillance
in the sense that smog represents more and more frequently a source
of trouble to outdoor imaging and brings about, among other problems,
low brightness and poor contrast, which greatly affect subsequent
tasks of image segmentation, tracking, target detection, etc. Extensive
research into this subject has been conducted by foreign and Chinese
scholars,with a range of achievements, both theoretical and practical,
resulted.\citep{choi2015referenceless,zhu2015fast,zhao2015single,wang2014single,tang2014robust,tang2016robust,zhang2015research}

Image defogging methods fall into two broad categories\citep{wudi2015review}:
image enhancement-based enhancement, and physical model-based restoration.
Image enhancement method, which does not account for the causes of
image degradation, is effective in improving contrast of images taken
on foggy days and bringing out their details, but entails some loss
of image information. Image restoration method, on the other hand,
tackles the problem by building a physical degradation model having
regard to the image degradation process on foggy days. This method,
attempting to invert the degradation process to arrive at fog-free
images, produces clearer defogging effect and involves less information
loss than image enhancement does, and so it is becoming a hot spot
in the community of image defogging research.

As for restoration method, a physical model in general use is the
atmosphere scattering model developed by Narasimhan et al.\citep{narasimhan2003contrast,narasimhan2002vision,schechner2001instant,cozman1997depth,nayar1999vision}.
With this model, it is possible to get satisfactory restoration results
by merely determining the transmissivity and the atmosphere light
value of each point in the image. It is, however, not an easy job
trying to determine these values. Okaley et al.\citep{oakley1998improving}
rely on radar installation to acquire the scene depth in order to
establish the transmissivity function; Narasimhan et al.\citep{narasimhan2003contrast}
extract the scene depth by use of two images taken under different
weather conditions; Nayar et al.\citep{narasimhan2003interactive}
seek to determine the scene depth and atmosphere light value by manually
specifying the sky zone in the image. All these methods are too often
not helpful in practical use because of certain restrictions. Over
recent years, He came up with an algorithm\citep{he2011single}: dark
original color a prior defogging, which attracts broad attention for
its simplicity and efficacy. Dark original color a priori knowledge
enables quick acquiring from the original image the transmissivity
corresponding to each point, thus making real-time defogging possible,
a feature at premium for outdoor surveillance. Its practical application,
however, usually produces results that are troubled by color oversaturation,
leading to image distortion. One significant reason is that the atmosphere
light value is estimated too roughly. He's estimation of atmosphere
light is largely based on the dark channel image acquired from dark
original color a priori. It works by selecting in the dark channel
image some brighter points as candidate atmosphere light locations
and then taking as atmosphere light points the brightest points of
the pixels, in the original image, corresponding to these locations.
This technique works well when the image contains a substantial area
of sky, but may give rise to gross discrepancies in locating atmosphere
light when the image contains no such sky area or when the image includes
other large, bright, and white objects, as illustrated in Fig.\ref{Fig:Orignal_Location_Hongkong},
\ref{Fig:Orignal_Location_Train}, and \ref{Fig:Orignal_Location_Swan}.

To cope with this problem, this paper presents an atmosphere light
estimation method that is based on light source point cluster selection.
The idea behind this method is performing clustering analysis of potential
atmosphere light points to find out all possible atmosphere light
zones, and then locating the atmosphere light by use of statistical
information of potential points in each zone.

The rest of this paper is organized as follows. Section \ref{sec:1}
gives an overview of atmosphere scattering model and dark original
color a priori defogging algorithm and analyzes the drawbacks, and
also their reasons, of the algorithm. In consideration of what has
been described, Section \ref{sec:2} proposes an atmosphere light
estimation method that is based on light source point cluster selection.
Section \ref{sec:3} looks into a comparative experiment and substantiates
the efficacy of the proposed algorithm from several indicators including
atmosphere light location and image quality. Also in this section,
the processing times of the algorithms are reported, confirming the
feasibility of the proposed algorithm in actual applications.

\section{Atmosphere Scattering Model\label{sec:1}}

Narasimhan et al.\citep{narasimhan2003contrast,narasimhan2002vision,schechner2001instant,cozman1997depth,nayar1999vision}
develop an atmosphere scattering model as below that describes the
degradation process of images:
\begin{equation}
I(x)=J(x)t(x)+A(1-t(x))\label{eq:1}
\end{equation}

where, $I$ is the intensity of the surveillance image, $J$ the light
intensity of the scene, $A$ the \textbf{atmosphere light} at infinity,
and $t$ the \textbf{transmissivity}. The first term, $J(x)t(x)$
, is called decay term, and the second term, $A\left(1-t\left(x\right)\right)$,
is called atmosphere light term. The aim of defogging is to restore
$J$ from out of $I$. To solve Eq.(\ref{eq:1}), we must determine
$t(x)$ and $A$ in the first place.

\subsection{Dark original color a priori knowledge}

Dark original color a priori comes from statistics derived from outdoor
images taken on fog-free days, and it gives us a piece of a priori
knowledge: in most outdoor fog-free images there are always some pixels
having a minimal light value in a certain color channel. This minimal
light value is termed \textbf{dark original color}, and the corresponding
pixels are known as dark original color pixels. The dark original
color a priori knowledge may be defined by:

Let:

\[
J^{dark}\left(x\right)=\underset{c\in\left\{ r,g,b\right\} }{\min}\left(\underset{y\in\Omega\left(x\right)}{\min}\left(J^{c}\left(y\right)\right)\right)
\]

where, $c$ represents a certain color channel, $J^{c}$ the component
of image $J$ in this channel, and $\Omega(x)$ a square zone with
pixel $x$ as its center.

If $J$ is an outdoor fog-free image, then for a certain pixel $x$
there exist always some pixels in its neighborhood $\Omega(x)$ such
that $J^{dark}(x)\approx0$, or:

\begin{equation}
J^{dark}\left(x\right)=\underset{c\in\left\{ r,g,b\right\} }{\min}\left(\underset{y\in\Omega\left(x\right)}{\min}\left(J^{c}\left(y\right)\right)\right)\text{\ensuremath{\approx}0}\label{eq:2}
\end{equation}

We now call $J^{dark}(x)$ \textbf{dark original color}, and refer
to the above rule as \textbf{dark original color a priori knowledge}.
This priori knowledge suggests that with fog-free images the $J^{dark}$
value is invariably minimal or close to 0.

\subsection{Image defogging method based on dark original color a priori}

\subsubsection{Estimation method of transmissivity}

In images tainted by fog, the intensity of $J^{dark}$ is higher than
usual because of the superposition of white light component in the
atmosphere, and for this reason such dark original colors are useful
in estimating roughly the transmissivity $t(x)$ in foggy images.

Suppose the atmosphere light $A$ is given, and the transmissivity
in a certain local neighborhood $\Omega(x)$ is constant. Apply the
minimum operator to Eq.(\ref{eq:1}), then divide the result by $A$
to get:

\[
\underset{y\in\Omega\left(x\right)}{\min}\left(\frac{I^{c}\left(y\right)}{A^{c}}\right)=\tilde{t}\left(x\right)\underset{y\in\Omega\left(x\right)}{\min}\left(\frac{J^{c}\left(y\right)}{A^{c}}\right)+\left(1-\tilde{t}\left(x\right)\right)
\]
where, $c$ represents the color channel, $\tilde{t}(x)$ tells us
that this is a rough estimate only.

Applying the minimum operator to color channel $c$ gives:

\begin{equation}
\underset{c}{\min}\left(\underset{y\in\Omega\left(x\right)}{\min}\left(\frac{I^{c}\left(y\right)}{A^{c}}\right)\right)=\tilde{t}\left(x\right)\underset{c}{\min}\left(\underset{y\in\Omega\left(x\right)}{\min}\left(\frac{J^{c}\left(y\right)}{A^{c}}\right)\right)+\left(1-\tilde{t}\left(x\right)\right)\label{eq:3}
\end{equation}

But dark original color a priori knowledge says that in the case of
fog-free outdoor images $J^{dark}$ should be close to 0. Substituting
Eq.(\ref{eq:2}) into Eq.(\ref{eq:3}) gives a rough estimate of the
transmissivity:

\begin{equation}
\tilde{t}\left(x\right)=1-\underset{c}{\min}\left(\underset{y\in\Omega\left(x\right)}{\min}\left(\frac{I^{c}\left(y\right)}{A^{c}}\right)\right)\label{eq:5}
\end{equation}

It is interesting to note that images would look unnatural and the
sense of depth is lost if the fog is removed completely. It is therefore
necessary to introduce into Eq.(\ref{eq:5}) a constant $\omega\left(0<\omega\le1\right)$
($\omega$ usually takes a value of 0.95) in order to keep some amount
of foggy effect:

\begin{equation}
\tilde{t}\left(x\right)=1-\underset{c}{\omega\min}\left(\underset{y\in\Omega\left(x\right)}{\min}\left(\frac{I^{c}\left(y\right)}{A^{c}}\right)\right)\label{eq:6}
\end{equation}

The transmissivity given by the above equation is rather rough, and
some block effect would be brought into the results. The algorithm
in question uses an image matting algorithm\citep{levin2008closed}
to get refined transmissivity $t(x)$. It is attained by solving the
following linear equation:

\begin{equation}
\left(L+\lambda U\right)t=\lambda\tilde{t}\label{eq:7}
\end{equation}

where, $\lambda$ is the correction factor, $L$ the Laplace matrix
suggested by the matting algorithm, typically a large sparse matrix.

\subsubsection{Estimation method of atmosphere light}

The image defogging method based on dark original color a priori estimates
atmosphere light $A$ by: organizing $J^{dark}$ in a descending order
of brightness value, taking the top $0.1\%$ pixels as candidate atmosphere
light points, comparing the brightness values of their corresponding
points in the original image $I$, and taking the pixel with the highest
brightness as $A$. This approach allows quick while accurate estimation
of the atmosphere light location.

\subsubsection{Image restoration}

With refined transmissivity $t(x)$ and atmosphere light $A$ known,
the restored image $J(x)$ may be found by the following equation:

\begin{equation}
J\left(x\right)=\frac{I\left(x\right)-A}{\max\left(t\left(x\right),t_{0}\right)}+A\label{eq:8}
\end{equation}
where, $t_{0}$ is a threshold designed to avoid too small a denominator,
and it is typically set equal to 0.1.

\subsubsection{Problems}

Practical application reveals that this method sometimes sets atmosphere
light $A$ on the brightest pixels, a place far off from real atmosphere
light, rather than in a zone where the fog is the heaviest, for instance
on white buildings or other large patches of white object. This observation
is illustrated in Fig.\ref{Fig:Orignal_Location_Hongkong}, \ref{Fig:Orignal_Location_Train},
and \ref{Fig:Orignal_Location_Swan}.

The method in discussion ranks $J^{dark}$ by brightness value and
chooses a brighter but not the brightest pixel as candidate atmosphere
light. This indeed precludes the brightest point from being chosen
as atmosphere light point, but has its own problem nevertheless: $J^{dark}$
value is \textit{$\Omega\left(x\right)$}-dependent, in other words
it is influenced by \textit{$\Omega\left(x\right)$} size. This method
would put the estimated atmosphere light on a bright object if this
object happens to be larger than \textit{$\Omega\left(x\right)$}in
size. In Fig.\ref{Fig:Orignal_Location_Hongkong}, we can see that
this method mistakenly puts the atmosphere light on the white building
and, in Fig.\ref{Fig:Orignal_Location_Train} and \ref{Fig:Orignal_Location_Swan},
the atmosphere light is mistakenly put on the headlight and the white
swan.

To cope with this problem, this paper introduces an atmosphere light
estimation method that is based on light source point cluster selection.

\section{Atmosphere Light Estimation Method Based on Light Source Point Cluster
Selection\label{sec:2}}

With an eye to solving the problems associated with the above-described
method, this paper comes up with an estimation method based on light
source point cluster selection, which works by acquiring, by use of
the dark original color a priori knowledge, the dark original color
$J^{dark}\left(x\right)$ corresponding to image $J(x)$, ranking
pixels $x$ by their brightness value, selecting the top $0.1\%$
pixels to form a set $X_{C}$ of candidate atmosphere light points:
\[
X_{C}=\left\{ x_{cand}|J^{dark}\left(x_{cand}\right)\in High\_Value\_Range\right\}
\]
and then performing statistics clustering on point set $X_{C}$. This
clustering is performed using the popular Euclidean distance $K-Means$
algorithm, with the clustering constant $K$ taking a value of 5:
\[
L=\{L_{n}|L_{i}\bigcap_{i\neq j}L_{j}=\varnothing;\bigcup_{n=1}^{5}L_{n}=X_{C};i,j,n=1,2,\cdots,5\}
\]
Thus the task is to find a point set such that the following equation
is minimized:

\[
J=\sum_{n=1}^{K=5}\sum_{x_{i}\in L_{n}}\|x_{i}-\mu_{n}\|^{2}
\]
where, $x_{i}$ is a point from point cluster $L_{n}$, and $\mu_{n}$
the geometric center of point cluster $L_{n}$ .

When the clustering result, $L=\{L_{n}|n=1,2,\cdots,5\}$, is found,
the point clusters $L_{n}$ are then ranked by the number $N_{n}$
of candidate light source points they contain. The point cluster with
the highest point number, designated by $L_{n'}$, is retained and
the zone from which it comes is taken as the \textquotedbl{}\textit{sky
zone}\textquotedbl{}, i.e., the zone of atmosphere light. Then, the
geometric center, designated by $\mu_{n}$ , of all the candidate
light source points $\{x_{i}^{'}|x_{i}^{'}\in L_{n'}\}$ in $L_{n'}$
is accepted as the location of atmosphere light, $(L_{\infty}^{R},L_{\infty}^{C})$,
in the image, or:

\begin{eqnarray*}
L_{\infty}^{R} & = & \frac{\sum\limits _{i=1}^{N_{n'}}Row\_Idx\left(x_{i}^{'}\right)}{N_{n'}}\\
L_{\infty}^{C} & = & \frac{\sum\limits _{i=1}^{N_{n'}}Col\_Idx\left(x_{i}^{'}\right)}{N_{n'}}
\end{eqnarray*}
The mean of the brightness vectors of all the candidate light source
points $\{x_{i}^{'}|x_{i}^{'}\in L_{n'}\}$ in $L_{n'}$ is taken
as the atmosphere light brightness vector $L_{\infty}$. Or:

\[
L_{\infty}=mean\left(J\left(x_{i}^{'}\right)\right)
\]

Now that the atmosphere light brightness vector $L_{\infty}$ has
been estimated, Eq.(\ref{eq:6}) enables us to determine the transmissivity
$t(x)$, and finally Eq.(\ref{eq:8}) serves to find the defogging
effect.

\section{Experiment and Analysis\label{sec:3}}

\begin{figure}[h]
\hfill{}\subfloat[Hongkong]{\begin{centering}
\includegraphics[width=0.4\textwidth]{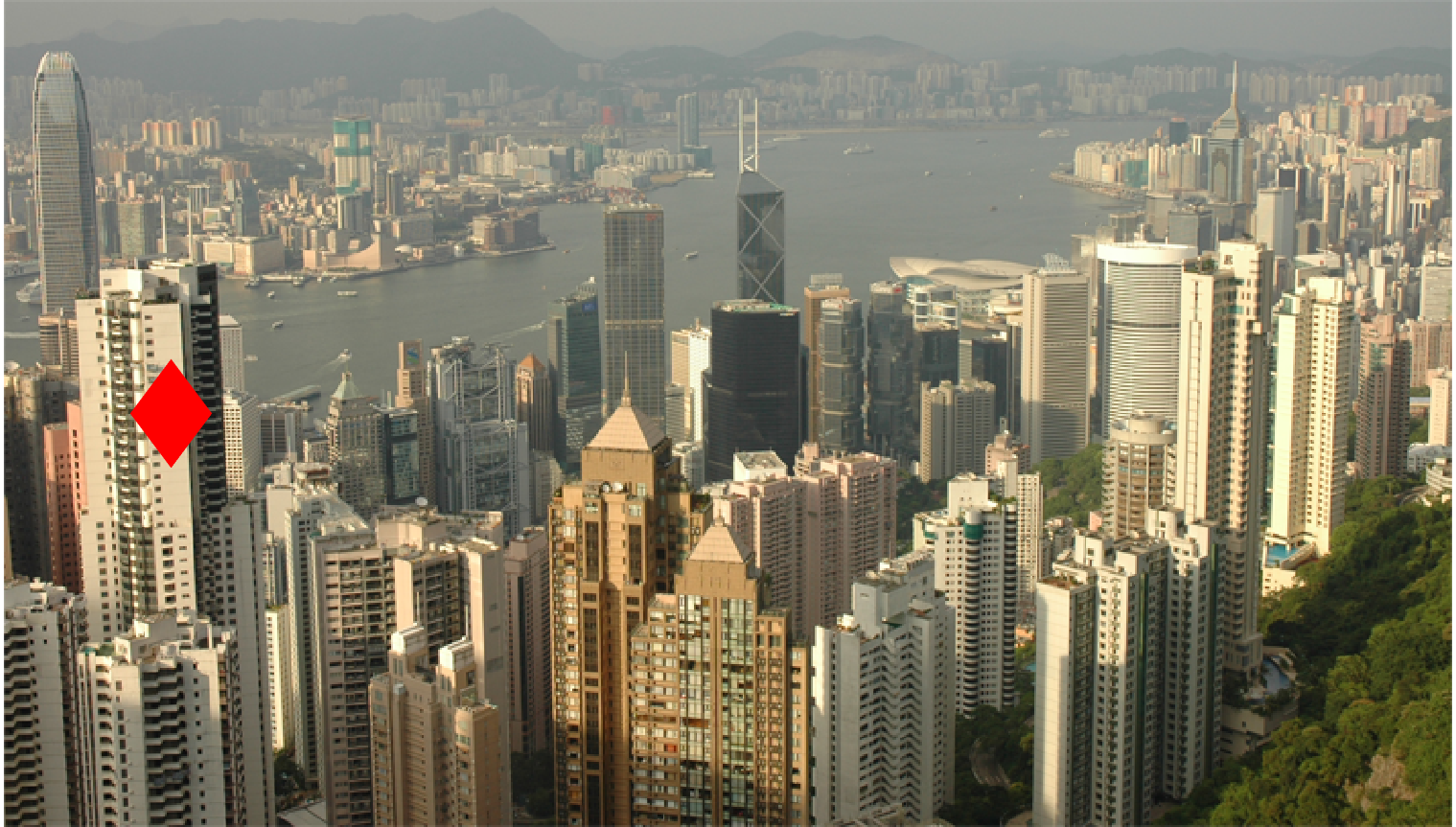}
\par\end{centering}

\label{Fig:Orignal_Location_Hongkong}}\hfill{}\subfloat[Hongkong]{\begin{centering}
\includegraphics[width=0.4\textwidth]{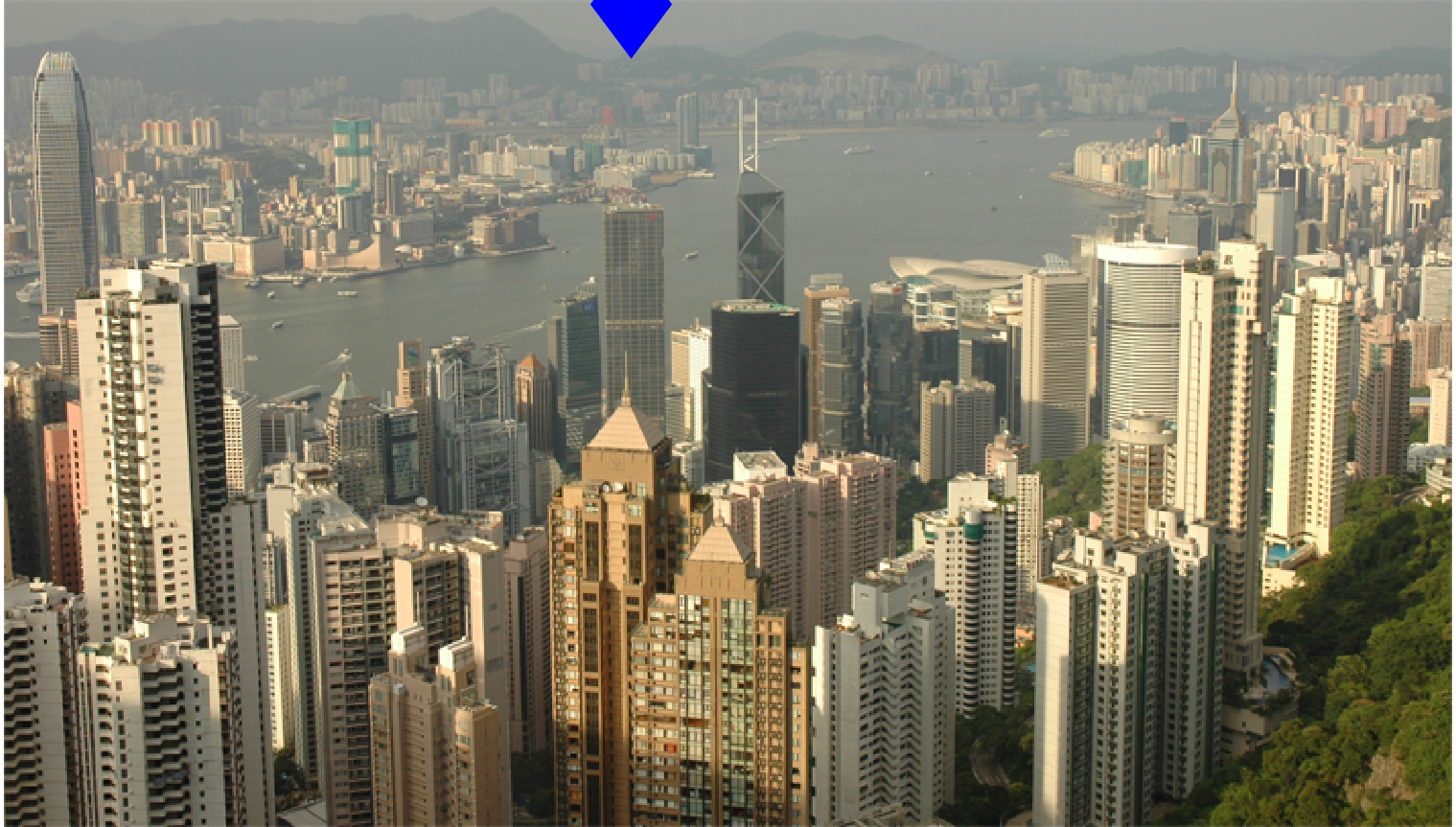}
\par\end{centering}

\label{Fig:Propose_Location_Hongkong}}\hfill{}

\hfill{}\subfloat[Train]{\begin{centering}
\includegraphics[width=0.4\textwidth]{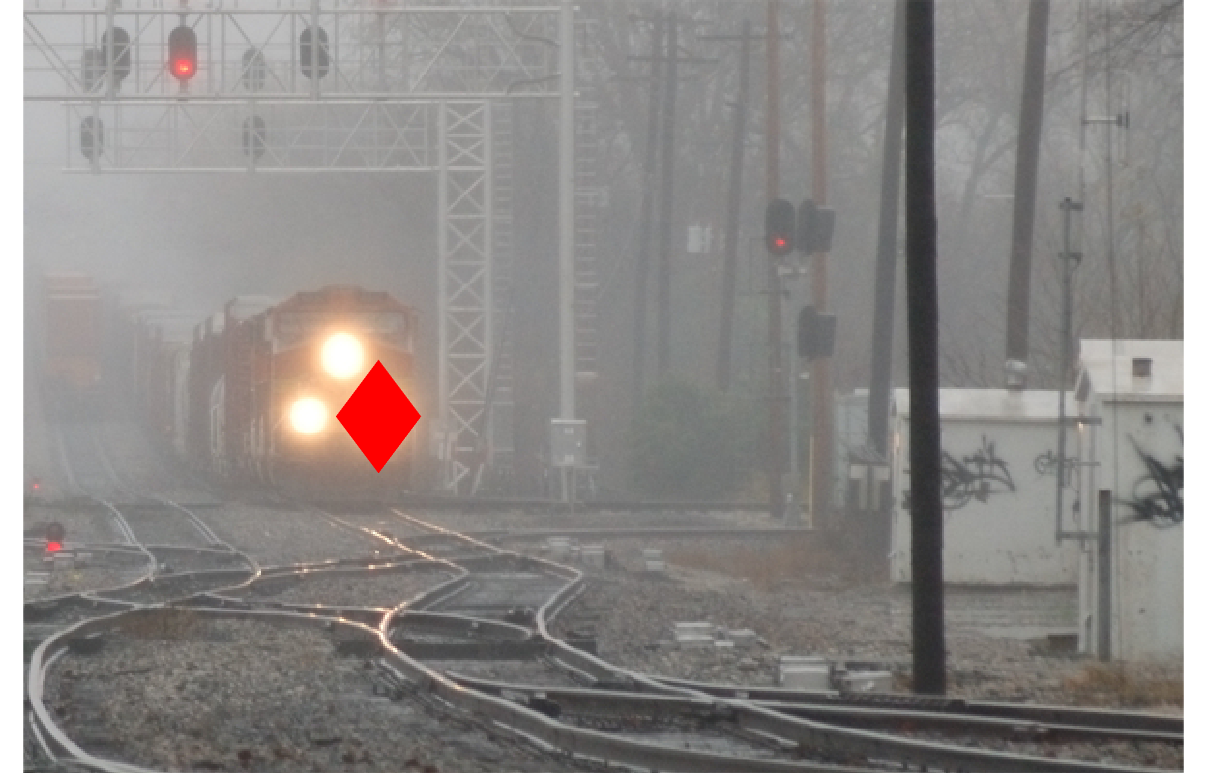}
\par\end{centering}

\label{Fig:Orignal_Location_Train}}\hfill{}\subfloat[Train]{\begin{centering}
\includegraphics[width=0.4\textwidth]{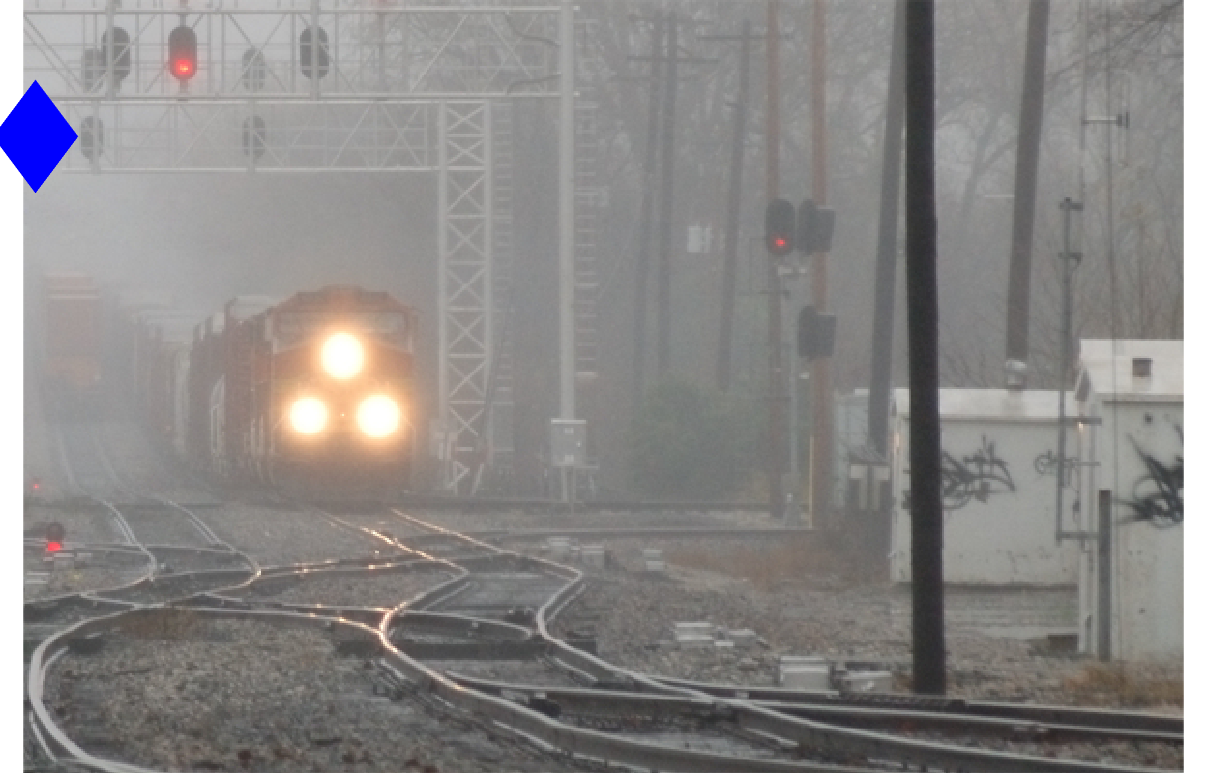}
\par\end{centering}

\label{Fig:Propose_Location_Train}}\hfill{}

\hfill{}\subfloat[Swan]{\begin{centering}
\includegraphics[width=0.4\textwidth]{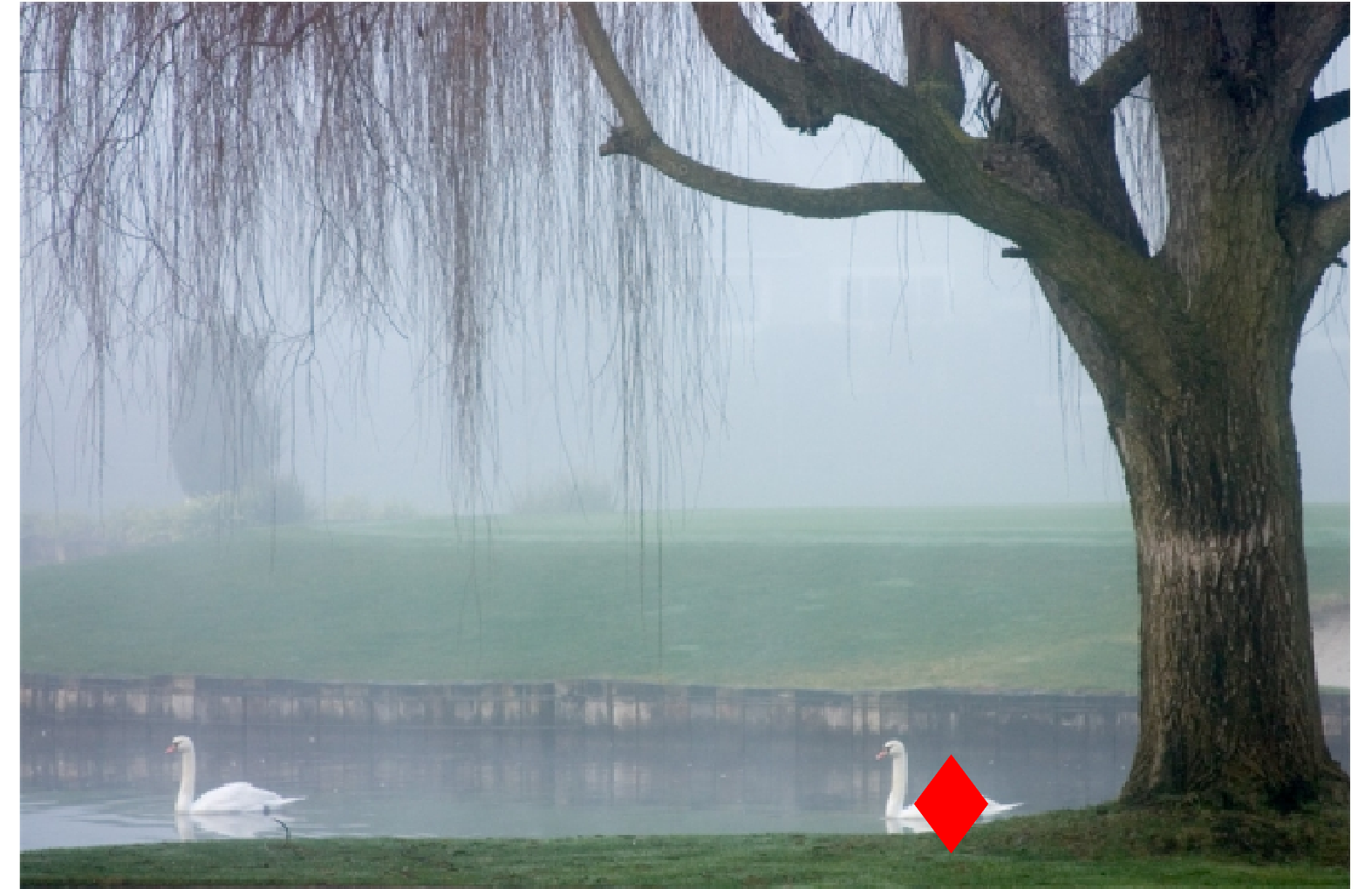}
\par\end{centering}

\label{Fig:Orignal_Location_Swan}}\hfill{}\subfloat[Swan]{\begin{centering}
\includegraphics[width=0.4\textwidth]{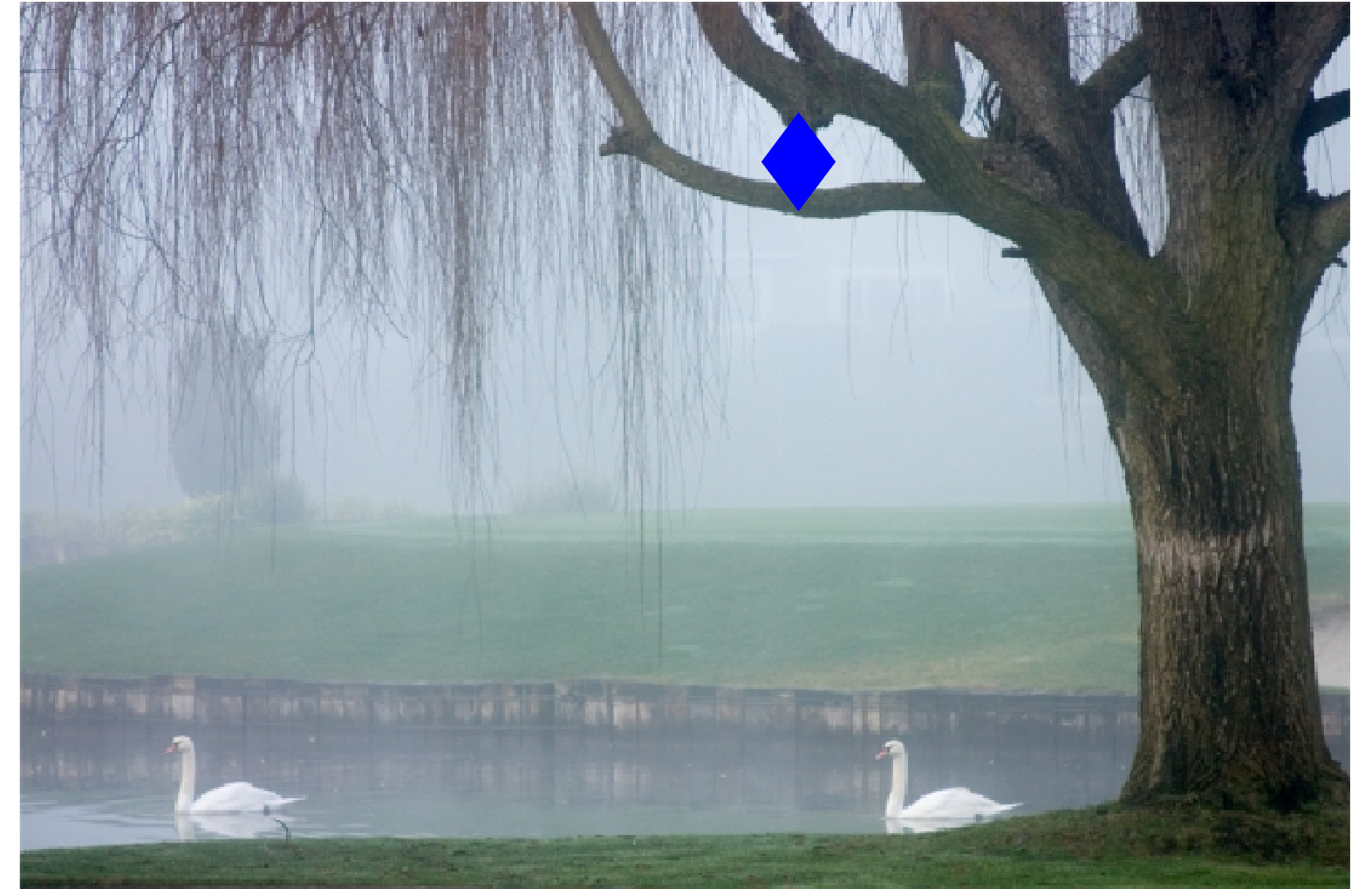}
\par\end{centering}

\label{Fig:Propose_Location_Swan}}\hfill{}

\caption{A comparison of atmosphere light location estimation: The left is
the estimation by the old algorithm (red diamond symbol) and the right
is the estimation by the proposed algorithm (blue diamond symbol).}

\label{Fig:Atmo_Location_Compare}
\end{figure}

\begin{figure}
\begin{centering}
\hfill{}\subfloat[Original transmissivity.]{\begin{centering}
\includegraphics[width=0.3\textwidth]{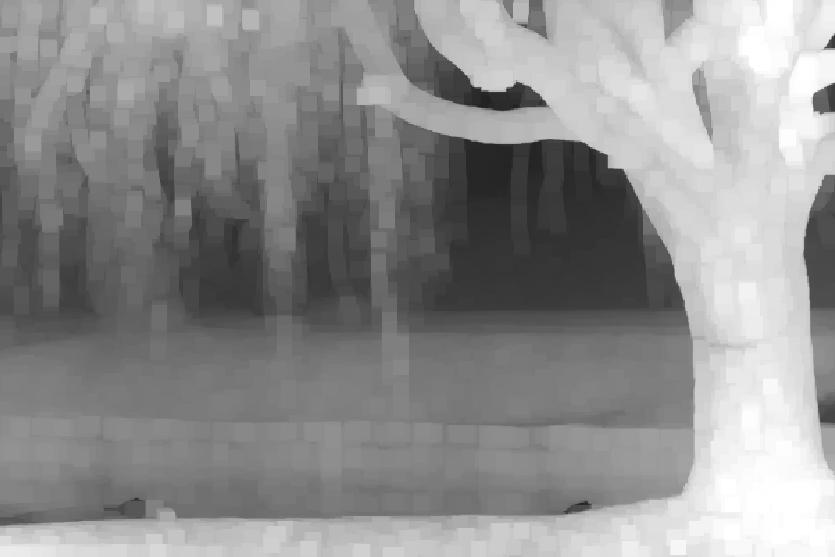}
\par\end{centering}
}\hfill{}\subfloat[Refined transmissivity by image matting algorithm.]{\centering{}\includegraphics[width=0.3\textwidth]{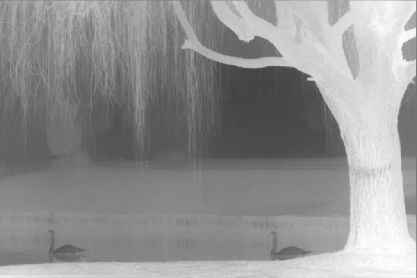}}\hfill{}\subfloat[Refined transmissivity by engineering method.]{\begin{centering}
\includegraphics[width=0.3\textwidth]{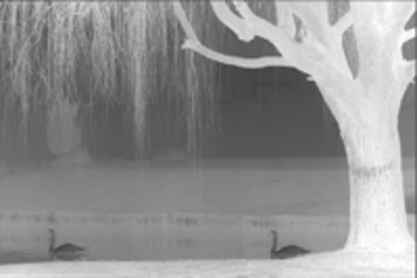}
\par\end{centering}
}\hfill{}
\par\end{centering}
\caption{Comparison of Refined Transmissivities.}

\label{Fig:Transmission_Compare}
\end{figure}

\begin{figure}
\begin{centering}
\subfloat[Forest]{\begin{centering}
\includegraphics[width=0.8\textwidth]{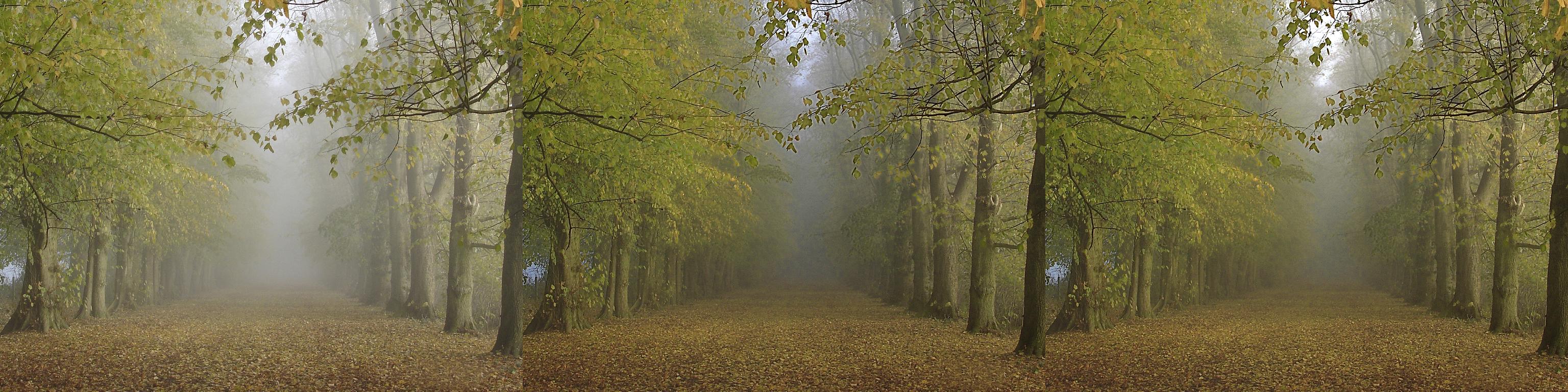}
\par\end{centering}
}
\par\end{centering}
\begin{centering}
\subfloat[Cones]{\begin{centering}
\includegraphics[width=0.8\textwidth]{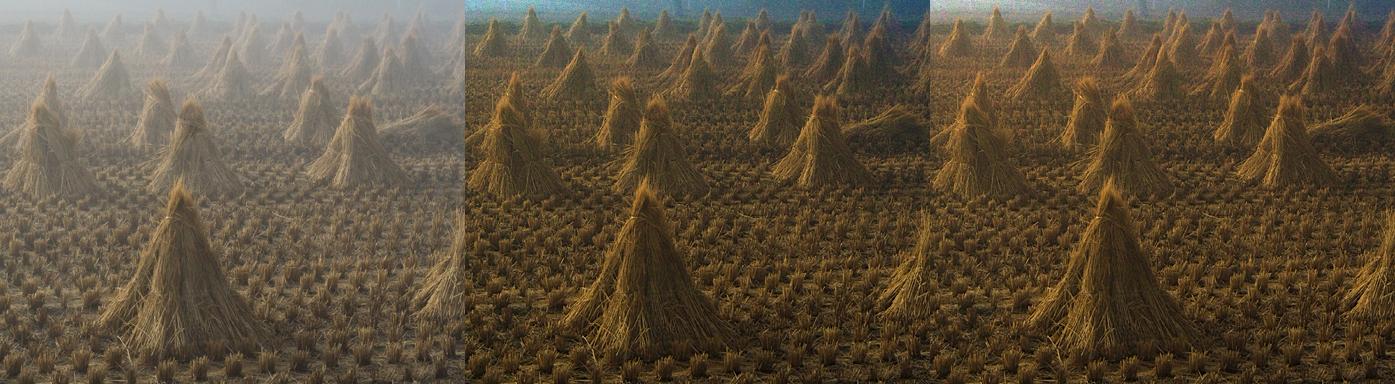}
\par\end{centering}
}
\par\end{centering}
\begin{centering}
\subfloat[Pumpkins]{\begin{centering}
\includegraphics[width=0.8\textwidth]{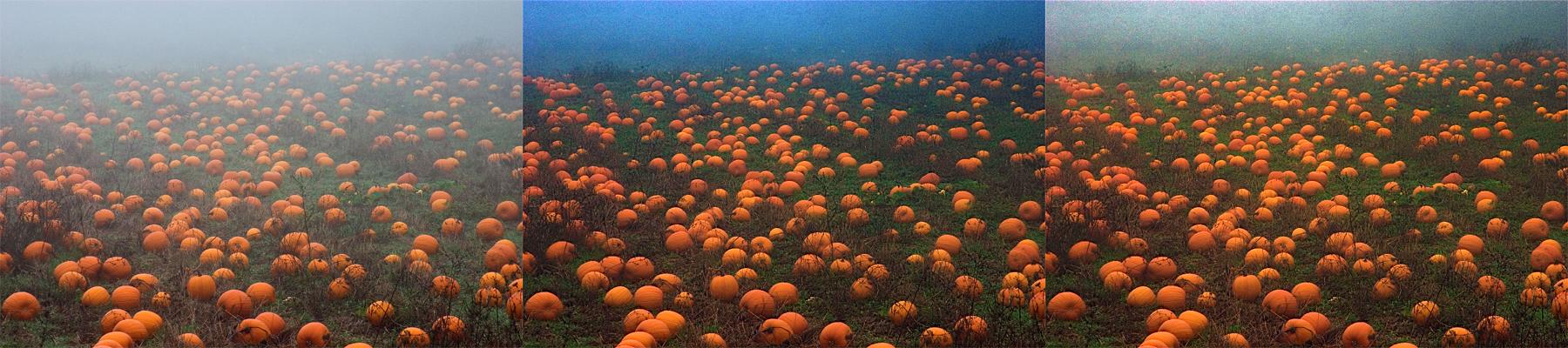}
\par\end{centering}
}
\par\end{centering}
\begin{centering}
\subfloat[Hongkong]{\begin{centering}
\includegraphics[width=0.8\textwidth]{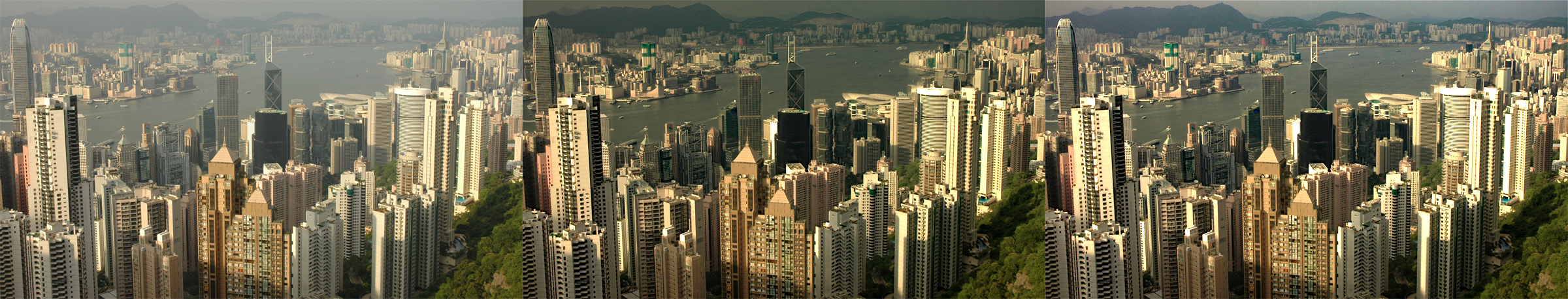}
\par\end{centering}
}
\par\end{centering}
\begin{centering}
\subfloat[Train]{\begin{centering}
\includegraphics[width=0.8\textwidth]{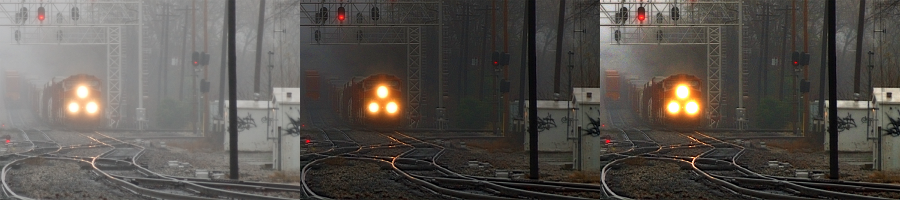}
\par\end{centering}
}
\par\end{centering}
\begin{centering}
\subfloat[Swan]{\begin{centering}
\includegraphics[width=0.8\textwidth]{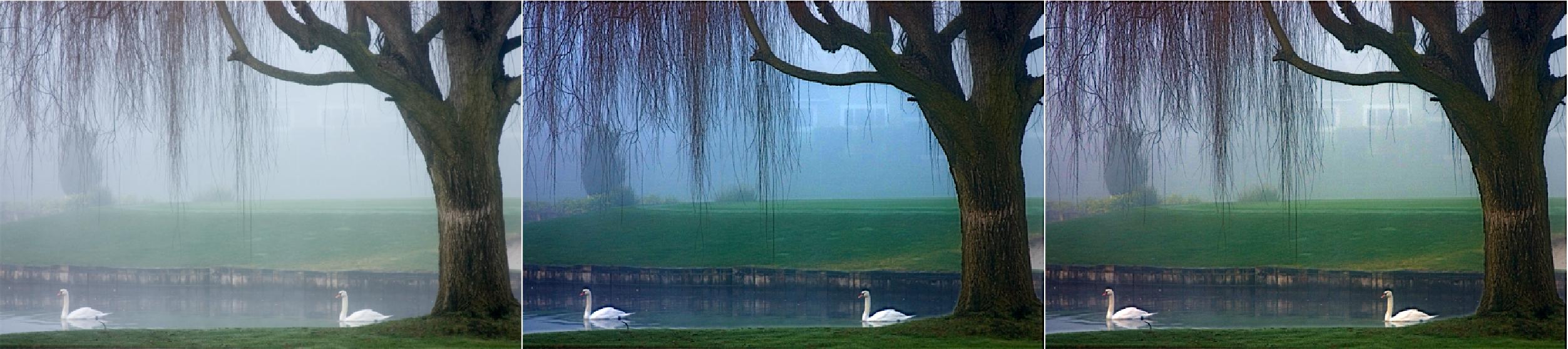}
\par\end{centering}
}
\par\end{centering}
\caption{Comparison of Defogging Result. From left to right: Original image,
result of original algorithm, result of proposed algorithm.}
\label{Fig:Dehaze_Result_Compare}
\end{figure}

In this study, the dark original color was found using a window size
$\Omega$ of $15\times15$, with Ω in Eq.(\ref{eq:6}) taking a value
of 0.95. Image matting was performed using the same algorithm as that
in Ref.\citep{levin2008closed}. The experiment hardware platform
was: Intel(R) Core(TM) i3-3220 @ 3.3GHz for CPU; 8GB DDR3 @ 1600MHz
for memory. The software platform was: MATLAB R2016a 64-bit. All the
algorithms were implemented using MATLAB code.

The old algorithm\citep{he2011single} relies on image matting to
refine the transmissivity, a technique that in essence amounts to
solving a large system of sparse linear equations, rather complex
in both space and time. But this step is designed to remove the block
effect arising from local windowing in estimating the transmissivity.
Therefore, this study turns to a technique commonly used in practical
engineering: first downscale the image to a suitable size, then refine
the transmissivity by image matting, and finally restore the reduced
image to its previous size using the refined transmissivity by means
of cubic interpolation. Fig.\ref{Fig:Transmission_Compare} compares
the transmissivities estimated using the two methods and we see that
they do not differ much in removing transmissivity-related block effect.

\subsection{Location of atmosphere light}

For comparison, the location estimated by the old method (denoted
by red diamond) and that estimated by the proposed method (denoted
by blue diamond) are shown in Fig.\ref{Fig:Atmo_Location_Compare}.
As shown in Fig.\ref{Fig:Orignal_Location_Hongkong}, the old algorithm
mistakes the building, which is bright and constitutes a large portion
of this image, for the atmosphere light; in contrast, the proposed
algorithm locates correctly the atmosphere light in the sky zone(Fig.\ref{Fig:Propose_Location_Hongkong}).
In Fig.\ref{Fig:Orignal_Location_Train}, the old algorithm makes
the mistake of locating atmosphere light on the headlight, the brightest
spot in the image, but the proposed algorithm accurately places atmosphere
light on the upper left sky zone, despite that it makes up only a
small portion of the image. Once more, in Fig.\ref{Fig:Orignal_Location_Swan}
the old algorithm commits the error of taking the white swan as atmosphere
light, but the proposed algorithm correctly recognizes the tree branch
zone, the portion with the heaviest fog, as atmosphere light.

\subsection{Defogging effect}

Fig.\ref{Fig:Dehaze_Result_Compare} provides a comparison of the
images restored from smoggy images by the two algorithms, which have
different atmosphere light estimation rationales. As is evident from
this figure, relative to the old algorithm the proposed one produces
effects that look more natural.

\subsection{Quantitative indicators}

In order to provide an objective evaluation of the defogging effect,
three indicators, i.e., contrast enhancement ratio, EME, and blind
assessment, are used to quantify the defogging effect.

Contrast is an important indicator of image quality because images
with good contrast typically look clearer and better and those with
a low contrast appear fuzzy or blurred. Table \ref{Table:Exp_Contrast_EME}
lists the contrast enhancement ratios, for both the old and the proposed
algorithms. It is obvious from the table that the proposed algorithm
performs better in terms of contrast enhancement.

EME, an image quality indicator developed by Agaian et al.\citep{agaian2000new},
is more consistent with human visual system (HVS), and here a higher
EME stands for better image quality. Table \ref{Table:Exp_Contrast_EME}
gives the EME values, again for both the old and the proposed algorithms.
From this table we see that both algorithms improve the EME substantially
in relation to the original images, and the proposed algorithm behaves
better than the old one in terms of defogging effect EME in all but
the swan image.

Blind assessment\citep{hautiere2008blind} is a quantitative indicator
developed by Hautiere et al. for evaluating the restoration quality
of images taken on foggy days. Its logarithmic image processing (LIP)
model assesses the defogging effect from three perspectives: the ratio
between new visible edges ($e$) before and after restoration (same
below), the ratio between mean gradient ($r$), and ratio between
black pixel proportion ($\sigma$). The higher the value of $e$ and
$r$ and the lower the value of $\sigma$, the better the defogging
effect after restoration. Table \ref{Table:Exp_Blind} provides a
comparison of different indicators associated with the two algorithms
and it appears that the proposed algorithm improves on the old one
in all these indicators.

\begin{table}[h]
\begin{centering}
\begin{tabular}{c>{\centering}p{0.1\textwidth}>{\centering}p{0.1\textwidth}c>{\centering}p{0.1\textwidth}>{\centering}p{0.1\textwidth}>{\centering}p{0.1\textwidth}}
\toprule
\multirow{2}{*}{Image} & \multicolumn{2}{c}{Contrast enhancement ratios} &  & \multicolumn{3}{c}{EME}\tabularnewline
\cmidrule{2-3} \cmidrule{5-7}
 & Original algorithm & Proposed algorithm &  & Original Image & Original algorithm & Proposed algorithm\tabularnewline
\midrule
Forest & 0.46 & 0.55 &  & 506.08 & 5207.98 & 5446.66\tabularnewline
\cmidrule{2-3} \cmidrule{5-7}
Cones & 0.94 & 0.98 &  & 0.98 & 999.58 & 1129.13\tabularnewline
\cmidrule{2-3} \cmidrule{5-7}
Pumpkins & 0.73 & 0.81 &  & 53.36 & 1611.85 & 2020.85\tabularnewline
\cmidrule{2-3} \cmidrule{5-7}
Hongkong & 0.65 & 0.96 &  & 21.99 & 5926.53 & 6365.53\tabularnewline
\cmidrule{2-3} \cmidrule{5-7}
Train & 0.98 & 2.85  &  & 0.47 & 2267.75 & 2798.29\tabularnewline
\cmidrule{2-3} \cmidrule{5-7}
Swan & 2.20 & 2.36 &  & 25.22  & 2639.50 & 2444.04\tabularnewline
\bottomrule
\end{tabular}
\par\end{centering}
\caption{Comparison of contrast enhancement ratios and EME indicator.}

\label{Table:Exp_Contrast_EME}
\end{table}

\begin{table}
\begin{centering}
\begin{tabular}{cccccccc}
\toprule
\multirow{2}{*}{Image} & \multicolumn{3}{c}{Original algorithm} &  & \multicolumn{3}{c}{Proposed algorithm}\tabularnewline
\cmidrule{2-4} \cmidrule{6-8}
 & $e$  & $r$ & $\sigma$ &  & $e$  & $r$ & $\sigma$\tabularnewline
\midrule
Forest & 0.111  & 1.198 & 0.001 &  & 0.111 & 1.229 & 0.001\tabularnewline
Cones & 0.324 & 1.528 & 0.000 &  & 0.326 & 1.547 & 0.000\tabularnewline
Pumpkins & 0.481 & 1.715 & 0.000 &  & 0.463 & 1.759 & 0.000\tabularnewline
Hongkong & 0.051 & 1.362 & 0.006 &  & 0.040 & 1.525 & 0.006\tabularnewline
Train & 1.553 & 1.450 & 0.007 &  & 1.497 & 1.968 & 0.007\tabularnewline
Swan & 0.517 & 1.705 & 0.001 &  & 0.499 & 1.736 & 0.002\tabularnewline
\bottomrule
\end{tabular}
\par\end{centering}
\caption{Blind indicator.}
\label{Table:Exp_Blind}
\end{table}

\subsection{Operational speed}

The proposed algorithm is designed to improve on images acquired by
outdoor surveillance instruments and thus its operational efficiency
represents an essential indicator as to its practical value. Table
\ref{Table:Exp_Speed} records the operation time of the proposed
algorithm and it may be concluded that this algorithm consumes minimal
time when running on mainstream outdoor surveillance hardware platforms.
For instance, it took only 0.108s when processing a forest image of
720P size (1080{*}720). It is useful to note that the tabulated results
were delivered by an algorithm implemented using Matlab; if implemented
in C/C++ or other similar languages, the algorithm is expected to,
according to experience, register a minimum of tenfold improvement
in computational efficiency, implying that a minimal load is created
to the legacy outdoor surveillance equipment. Moreover, parallel computation
typical of clustering makes it possible, by implementing an algorithm
of parallel version, to bring additional improvement to computational
efficiency.

\begin{table}
\begin{centering}
\begin{tabular}{ccc}
\toprule
Image & Size & Time consumed (s)\tabularnewline
\midrule
Forest & 768{*}1024 & 0.108\tabularnewline
Cones & 384{*}465 & 0.172\tabularnewline
Pumpkins & 400{*}600 & 0.030\tabularnewline
Hongkong & 457{*}800 & 0.066\tabularnewline
Train & 400{*}600 & 0.029\tabularnewline
Swan & 557{*}835 & 0.056\tabularnewline
\bottomrule
\end{tabular}
\par\end{centering}
\caption{Time consumed by the proposed algorithm (Matlab edition) for atmosphere
light estimation}

\label{Table:Exp_Speed}
\end{table}

\section{Conclusion}

This paper gives a theoretical analysis of the methodology used by
an existing defogging algorithm for acquiring atmosphere light and,
assisted by experimental observation, some drawbacks have been identified
in this methodology, and then the reasons of these drawbacks are investigated.
A technique for determining atmosphere light is subsequently proposed
that is based on light source point cluster selection.

Experimental results indicated that the proposed algorithm produces
more accurate information with respect to atmosphere brightness vector
and location (Fig.\ref{Fig:Atmo_Location_Compare}), the images restored
using this method look more natural subjectively (Fig.\ref{Fig:Dehaze_Result_Compare}),
and the objective image quality indicators are better (Table \ref{Table:Exp_Contrast_EME}
and \ref{Table:Exp_Blind}). Besides, the proposed algorithm consumes
less time (Table \ref{Table:Exp_Speed}), virtually giving no extra
load to legacy outdoor surveillance systems.

The future effort of this study is on how to improve the algorithm
such that its computational load is reduced and that it may be used
with on-board surveillance systems and on other practical occasions.

\bibliographystyle{gbt7714-2005}
\bibliography{References}

\end{document}